# Application of Rényi and Tsallis Entropies to Topic Modeling Optimization

**Koltcov Sergei,** *National Research University Higher School of Economics, St. Petersburg, 190008 Russia*

e-mail: skoltsov@hse.ru

**Abstract**:

## 1. Introduction

Statistical physics is increasingly being used to describe objects and processes that go beyond physical phenomena. Thus, large arrays of textual data, which have been rapidly accumulating on the Internet in the last decade, require ever more complex methods for their automatic processing and modeling. For this, a wide range of mathematical tools, including topic models, is used [1], but their properties and behavior remain little studied so far, which makes it impossible to choose the optimal parameters of such models. If, however, we consider the results of topic modeling as nonequilibrium complex systems (since these, as will be shown below, have the characteristics of such systems), this would make it possible to apply to them a whole range of approaches from statistical physics. First of all, these are models for analyzing the processes of self-organization of large ensembles. The basis for such an analysis may be an approach in which the behavior of the topic model of a textual collection as a word ensemble would be determined by thermodynamic functions, such as entropy or free energy. It is known that complex systems can be characterized by exponential and power law distributions, which is especially characteristic for social [2, 3], biological [4, 5] and economic systems [6, 7]. However, for topic models of textual collections, where the units are documents, words and latent semantic variables (topics), the Pareto-like distribution is characteristic [8, 9]. Proceeding from this, when applying the maximum entropy principle for such systems, an approach based on deformed statistic with the underlying Rényi or Tsallis entropy should be used [10, 11]. In this case, the deformed statistic of complex systems, like non deformed statistic, will describe the probabilistic regularities characterizing the topic model of the textual collection as a system of a large number of "particles", a system which, in turn, can remain in thermodynamically equilibrium and nonequilibrium states. Inclusion of the deformation parameter q in thermodynamically abnormal systems with long-range interactions will make it possible to explain the particularities of the behavior of such complex systems. Moreover, the search for optimal parameters describing the state of these systems can be achieved on the basis of the procedure for seeking the maximum entropy [12].

Our attention in this work is focused on topic modeling [1], since it is the most effective and sometimes the only available method of obtaining knowledge about the topic structure of large textual collections about which nothing is known in advance. This task is often encountered in the study of Internet content, including news, consumer reviews, and social network messages. In this case, topic modeling as a mathematical approach is applicable for the analysis not only of textual data [13], but also of mass spectra [14], images [15], and other objects. In essence, topic modeling is an expanded version of cluster analysis, which makes it possible to calculate simultaneously the distribution of words and documents by topics/clusters. Moreover, topic models also provide the opportunity to rank words and documents according to the probability within each topic/cluster, which is not typical for traditional cluster analysis. The major problem of this group of methods is the lack of ground truth, that is, of an idea of the correct number and composition of clusters, which inhibits the study of the properties of these models and forces us to seek approaches based on theories from other areas of knowledge.

Thus, in this paper we study the behavior of topic models with a change in the number of topics based on thermodynamic concepts and deformed entropy. The purpose of such a study is to find the optimal number/clusters of topics in the topic modeling, first, based on the maximum information approach, and second, on the basis of the T-invariance principle introduced in the work.

The work consists of the following parts. Section 2 first briefly explains the essence of topic modeling, which is necessary to further describe the proposed solutions. The second part of this section provides an overview of the available approaches for determining the optimal number of clusters in cluster analysis and topics in topic modeling, and their limitations are indicated. In Section 3, we propose our entropy approach to the analysis of topic models as complex nonequilibrium systems, on the basis of which it is possible to find the optimal number of topics. Sections 4 and 5 describe the data used and the results of numerical experiments to verify our approach. Section 4 shows that the minimum q-deformed entropy is reached with the 'correct' number of topics taken from the marked up textual data, and therefore can be used as a criterion for selecting the number of topics. Section 5 shows experimentally that the cumulative lexical composition of the topics is invariant, that is, it is resistant to changing the number of topics in the greater part of the range of variation, but this invariance gives quasiperiodic functions, which must be taken into account when selecting the number of topics. It is concluded that the T-invariance parameter must be included in the general theory of searching for optimal parameters of topic models, but it is beyond that article.

## 2. Problems of topic modeling and cluster analysis.

### 2.1. Introduction to topic modeling.

Topic modeling as a version of cluster modeling is based on the following provisions [16]:

1. Let D be a collection of text documents, and W, a set (dictionary) of all unique words. Each document d ∈ D is a set of terms $w_1,...,w_{nd}$ from dictionary W.

2. We assume that there is a finite number of topics T, and every entry of word w in document d is associated with some topic t ∈ T. A topic is taken to mean a set of words that are often found together in a large number of documents.

3. A collection of documents is considered a random and independent selection of triples $(w_i,d_i,t_i)$, i = 1,...,n from the discrete distribution p(w,d,t) on the finite probabilistic space W × D × T. Words w and documents d are observable variables, topic t ∈ T is a latent (hidden) variable.

4. It is assumed that the order of terms in documents is not important for identifying topics (the 'bag of words' approach). The order of documents in the collection also does not matter.

In TM (topic modeling), it is assumed that probability p(w|d) of the occurrence of terms w in documents d can be expressed by product distributions p(w|t) and p(t|d). According to the formula of total probability and the hypothesis of conditional independence, we have the following expression [17]:

$$p(w|d) = \sum_{t \in T} p(w|t)p(t|d) = \sum_{t \in T} \phi_{wt}\theta_{td} \qquad 1.$$

where p(w|t) is the distribution of words by topics, and p(t|d) is the distribution of documents by topics. Thus, to construct a topic data model means to solve the inverse problem in which it is necessary to find the set of hidden topics T, i.e., the set of one-dimensional conditional distributions p(w|t) ≡ φ(w,t) for each topic t that make up matrix $φ_{wt}$ (the word distribution by topics) and the set of one-dimensional distributions p(t|d) ≡ θ(t,d) (matrix $θ_{td}$, the document distribution by topics) for each document d based on the observed variables d and w.

Within the framework of topic modeling, two approaches to the inference of distributions are being actively developed: 1. TM working on the principle of maximum likelihood estimation. For the given approach, the most known models are probabilistic latent semantic analysis (pLSA) [17] and variational latent Dirichlet allocation (VLDA) [13], in which matrices p(w|t) and p(t|d) are found by EM-algorithm. 2. TM on the basis of Markov chain theory or a model with Gibbs sampling (LDA/GS) [18], wherein p(w|t) and p(t|d) are found by calculating the mathematical expectation by the Monte Carlo method. A brief description of the differences between these models is given in the supplementary material.

## 2.2. Nonequilibrium and instability of topic modeling.

The topic modeling process can be regarded as the transition of the entire system to a nonequilibrium state. The initial distribution of words and documents in matrices $\varphi_{wt}$ and $\theta_{td}$ with the use of LDA/GS is flat, and with the use of pLSA and VLDA is given by means of a random number generator. For both types of algorithms, the initial distribution corresponds to the maximum entropy. However, regardless of the type of algorithm and the initialization procedure, word and document probabilities in the topic models are redistributed by topics during modeling in such a way that a considerable proportion of word probabilities (about 95% of all unique words) is close to zero, and only about 3-5% have rather high probabilities [19, 20].

The similarity of solutions obtained in the course of topic modeling with the nonequilibrium state of physical systems makes it possible to apply concepts from statistical physics to the TM analysis, but this similarity is not absolute. Physical systems are characterized by the indistinguishability of particles: that is, if the topic solution was a physical system, it would not matter which particles (words) populated the states with high probability values. However, for textual content researchers, the specific composition of the most probable words of each topic and all the topics together is precisely the main informative result. This circumstance has two consequences.

First: the composition of each topic is important, but the nondeterministic nature of the TM algorithms leads to the fact that the same algorithm run on the same data with the same parameters yields slightly different topics. This does not allow the text researcher to answer the main question: what are the topics contained in the given collection? The problem of such TM instability is very little studied; one of the few solutions is the previously proposed extension for the LDA/GS algorithm - Granulated LDA (GLDA) [21, 20], which demonstrates almost 100% stability (for more details, see the supplementary material). Although GLDA has a number of serious shortcomings, such as weak interpretability and high topic correlation, we use this algorithm among others in our experiments in this study to test the applicability of the proposed approach to both low-deterministic and high-deterministic LDA algorithms.

Second: it is also important which words turn out to be most probable in all topics as a whole when changing the number of topics. If the topic solutions for different numbers of topics give radically different compositions of top words, the algorithm as a whole is not suitable for use. If, however, there are certain ranges of values of the number of topics in which the composition of top words is very different from most of the other solutions that are similar to each other, then cutting off such ranges, together with searching for the minimum entropy, can become an important criterion when selecting the optimal number of topics. This problem has not yet been investigated at all, and in this paper we are just approaching the first results of studying it. The degree of stability of the compositions of top words across models with different number of topics will be called T-invariance, where T is the number of topics/clusters.

**2.6. Approaches to T parameter determination**

The main problem in finding the optimal number of clusters in cluster analysis and topics in topic modeling is the choice of a function for optimization. In cluster analysis, minimization of intracluster distance and maximization of intercluster distance are most often used. The problem is that the increase in the number of clusters leads to quite a smooth monotonic dependence of this kind of function on the number of clusters. Accordingly, the development of transformation procedures is required to extract features from such functions [22], for which various solutions are proposed in the cluster analysis [23, 24, 25]. Clustering quality measures are discussed in works [26, 27]. There are also models for determining the number of clusters based on the entropy maximization principle [28, 29], but they use the Gibbs-Shannon entropy.

For our purposes, the most interesting approach is that of entropy employed in cluster analysis based on the ideas of statistical physics, namely, on the basis of minimizing the free energy [30]. Its main idea is as follows: each element of the system is characterized by probabilities of belonging to different clusters. Accordingly, for each element, we can formulate the concept of internal energy and calculate the free energy of the system. The temperature in such a system is considered as a free parameter, which varies in order to minimize the free energy. Such a thermodynamic approach is successfully used in the theory of dynamical systems [31] and in the analysis of images [32] and neural networks [33]. Further development of this approach occurs in the framework of nonextensive statistics. The discussion of the application of q-deformed statistic for machine learning is presented in [34, 35], and the generalized version of the 'rate distortion theory' is discussed in [36]. The possibility of applying deformed statistic for image segmentation is discussed in [37]. However, in none of these studies is q-deformed statistic used to determine the number of topics in topic modeling, although the problem of finding the optimal number of topics is just as relevant for it and even more complex.

This is due to the following reasons. Firstly, it is difficult in TM to formulate the semantic concept of a topic, the linguistic criteria for dividing two topics among themselves and, accordingly, the quality measures for topics and topic solutions. Secondly, in TM, as well as in cluster analysis, it is difficult to formulate an adequate functional dependence, which, on the one hand, would characterize the quality of the topic model, and on the other would be a function of the number of topics. Nevertheless, there are several works in which the authors have attempted to solve the problem of choosing the number of topics specifically in topic modeling. The authors of work [38], based on the ideas of cluster analysis, show that topic solutions with minimal correlation between the topics measured with the help of the cosine measure correspond to solutions with a minimum value of another quality measure, perplexity. This is an interesting work, but we have never come across data on which the function of the perplexity dependence on the number of topics would have a minimum (as it has as per the authors), instead of monotonically decreasing. In another work [39], the authors propose performing the singular-value decomposition (SVD) of matrices $\varphi_{wt}$ and $\theta_{td}$, then selecting two vectors containing singular quantities and calculating the distance between them on the basis of the Kullback-Leibler divergence, which it is proposed to minimize. In this case, the optimal number of topics corresponds to the situation where both matrices are described by the same number of singular quantities. The authors of this approach do not verify it with the help of alternative measures of the quality of the solutions obtained, and in addition the operations of SVD decomposition and calculation of the Kullback-Leibler divergence severely restrict the application of this approach to large data. The collections used in [39] did not exceed 2,500 texts.

One of the most well-known approaches to the problem of determining the number of topics in topic modeling is the 'Hierarchical Latent Dirichlet Allocation, (hLDA)' model [40, 41], which allows to construct a hierarchy of topics in the form of a tree with the initial assumption of the existence of an infinite number of topics. The choice of this or that branch, as well as the selection of the required number of levels in the tree, is determined by the specifics of the task, by the user, and by the features of the dataset. In addition, conditions are implicitly built inside the algorithm that limit the structure of the tree and, therefore, affect the total number of topics. Firstly, there is the parameter of concentration $\gamma$ that significantly affects the size of the tree [40], and secondly, there is the predefined constant that determines the number of topics that describe a single document [41]. These parameters must be set by the user on bases that are not completely clear, and their variation can lead to a change in the number of topics.

Finally, among the approaches to determining the number of topics, it is worth mentioning the principle of calculating the nonequilibrium free energy for determining the number of topics previously formulated by us in [42], but this was only for one TM algorithm with Gibbs sampling, and its work was demonstrated only on one dataset. In addition, the relationship of the number of topics with the composition of the most probable words was not investigated in [42].

In the framework of this paper, we propose an expansion of the thermodynamic approach for the analysis of topic models, constructed both on the basis of the EM-algorithm and on the basis of Gibbs sampling. The approach was expanded on the basis of recording Rényi and Tsallis entropies as the main quality measures necessary to find the optimal number of topics. The study of T-invariance of TM was realized using the Jaccard index.

**3. Application of the entropy approach to the analysis of complex textual systems.**

The number of topics can be considered as a parameter characterizing the 'algorithm resolution'. In this case, one should not study the topics separately, for example, using the cosine measure or SVD decomposition, but consider the set of topics as a statistical ensemble of words with high probability. Accordingly, it is possible to investigate the behavior of the distribution of the ensemble of words when the extensive parameter T, the 'number of topics', changes. In accordance with the maximum entropy principle [12], we consider entropy as negative information; thus, the maximum entropy corresponds to the minimum of information. We assume that the 'true number of topics' (the best resolution of the algorithm) corresponds to the maximum of the information received (or to the minimum of nonextensive entropy of the topic model) as a result of topic modeling.

Proceeding from this, the collection of documents can be considered as a mesoscopic information system consisting of millions of elements (words and documents) with an initially unknown number of topics. If we look at the change in the number of topics given by the researcher as a process in which the system exchanges information with the environment, then such a system will be an 'information thermostat' [11], which by definition, unlike a physical thermostat, is an open system. Accordingly, with a change in the number of topics, the information system may not reach an equilibrium state in the sense of the Gibbs-Shannon entropy maximum, but it may be in an intermediate equilibrium state, which is determined by the local minimum of Rényi or Tsallis entropy.

The totality of words that often (in the statistical sense) are found together in a large number of documents forms what can be called a topic. If similar topics are fairly consistently reproduced from solution to solution on the same datasets, then such a topic can be considered a dissipative structure in the sense of Prigogine [43]. A collection of documents can contain only a finite number of such structures. Therefore, the cumulative set of words with a probability above a certain threshold and, thus, characterizing all dissipative structures as a whole, presumably should be constant. It is these stable dissipative structures that should be identifiable through topic modeling.

As was partly mentioned in Section 2.2, the share of words with high probability is extremely small, and the very value of probabilities in top words differs sharply from that of all other words. That is, the distribution of words and documents in such information systems is extremely unbalanced, making it possible to determine the entropy of the system taking into account the nonequilibrium. Based on this, we will formulate our approach in the following provisions [42]: **1.** In the information thermodynamic system under consideration, the total number of words and documents is a constant, that is, there is no volume change. **2.** A topic is a state (analogous to the direction of spin) that each word and document can take in the collection. At the same time, both words and documents can belong to different topics with different probabilities. **3.** The information thermodynamic system is open and exchanges energy with the external environment by changing the temperature, which is understood as the number of topics (or clusters) T. This value is given from the outside and is a parameter that must be determined by searching for the minimum nonextensive entropy of the system. **4.** As a measure of the nonequilibrium of such an information system, one can use the entropy difference $\Lambda_s = S-S_0$ (Lyapunov function, relative entropy) [44], where $S_0$ is the entropy of the state, which is taken as the zero mark (chaos), and S is the entropy of the nonequilibrium state. By analogy with the Lyapunov entropy function, we can formulate a function constructed on the basis of the difference of free energies: $\Lambda_F = F(T)-F_0$, where $F_0$ is the free energy of the initial state (chaos), and F(T) the free energy at a given value of T [44]. **5.** Since the topic modeling algorithm is a procedure for restoring hidden distributions from a collection, the number of distributions is a variable parameter. The optimal number of such distributions corresponds to the situation where an information maximum is reached (minimum entropies). **6.** In this paper, it is also assumed that the equilibrium state of an information system can be characterized by the fact that a set of words with high probability ceases to change with a change in the number of topics. This means that the difference between two topic solutions, calculated using Jaccard index [45], can be a constant at a certain interval of the parameter 'number of topics'.

### 3.1. The density-of-states function

The total number of microstates in the information system that can be taken by words in the topics is equal to N*T, where N is the number of unique words in the collection of documents, and T is the number of topics/clusters. Let us define the density-of-states function as follows: $\rho(E) = \frac{\sum_{ij}^{NT} N(\varepsilon)_{nt}}{N \cdot T}$, where $N(\varepsilon)_{nt}$ - where $N(\varepsilon)_{nt}$ is the number of states with energy E for words with high probability, n,t is the summation over all words and topics. Accordingly, the relative Shannon entropy can be expressed in terms of the density of states as follows [33]:

$$S(E) = \ln(\rho(\varepsilon)) \qquad (2).$$

It should be noted that the sum of probabilities for all microstates is always a unity:

$$1 = \frac{1}{NT}\sum_{nt}^{NT} \rho(\varepsilon)_{nt} \qquad (3).$$

It should also be noted that relative Shannon entropy is an option of gap statistic [24].

### 3.2. The energy of the microstate and the statistical sum of the information system.

The energy of one microstate can be expressed as: $\varepsilon_{nt}=-\ln(p_{nt})$, where n is the word number in the list of unique words, t is the topic number, and $p_{nt}$ is the probability of word n in topic t. In general, the energy range can be divided by a given number of intervals k, therefore, the density-of-states function and energy can be written as:

$$\rho(E) = \frac{\sum_{nt}^{NT} N(\varepsilon)_{nt}}{NT} = \frac{\sum_{k}^{K} N_k}{NT}, \quad (4),$$

where $N_k$ is the number of microstates with energy $\varepsilon_k$ falling within interval k. The statistical sum can be written in the following form: $Z = \sum_{k}^{K} e^{-\varepsilon_k/T}$.

### 3.3. Nonequilibrium free energy of the topic model.

As already noted, the transition to a strongly nonequilibrium state occurs in the course of topic modeling, which is characterized by the fact that one part of the states has a high probability $P_{nt}>1/N$, and the other a low probability $P_{nt}<1/N$, close to zero. From here on, we will consider the states in which the information system resides with a non-zero probability. The entropy of a nonequilibrium system is described by the quantity $\Lambda_s = S-S_0$. Accordingly, the entropy and energy of the system are a function of the number of topics. Proceeding from the above, we can represent the nonequilibrium free energy of the topic model in the following form:

$$\Lambda_F = F(T) - F_0 = (E(T) - E_0) - (S(T) - S_0) \cdot T = -\ln\left(\frac{\sum_{t=1}^{T}\sum_{n=1}^{N} P_{nt}}{T}\right) - T \cdot \ln\left(\frac{N_{k1}}{N \cdot T}\right) \quad (5),$$

where $N_{k1}$ is the number of states in which $P_{nt}> 1/N$, (N·T) is the total number of all states, T is the number of topics (a variable parameter), N is the size of the dictionary of unique words, and $E_0 S_0$ is the energy and entropy of the system for the initial distribution that correspond to the maximum entropy. Quantities $N_{k1}$ and $P_{nt\ are}$ calculated for each topic model with variation of the parameter T, so the quantity $\Lambda_F$ is a function of T.

### 3.4. Information measure.

As mentioned above, since the information measure is represented as entropy taken with the reversed sign, that is, the maximum entropy corresponds to the minimum information [12], the search for the optimal number of topics in complex systems can be reduced to the search for the minimum entropy. The classical version of entropy is the Gibbs-Shannon entropy (Shannon entropy) [46]: $S = -\sum_i p_i \cdot \ln(p_i)$, which in the case of uniform distribution coincides with the Boltzmann entropy. In the framework of this paper, we also consider two main q-deformed entropies, Rényi and Tsallis, since they are suitable for analyzing the behavior of a nonequilibrium information system, and they are precisely the ones that we propose to minimize in order to find the optimal number of topics. The free energy of nonequilibrium information system $\Lambda_F$ is, on the one hand, $\Lambda_F = F = E - TS$, and, on the other hand, $F = \ln(Z)/T$. With this in mind, the statistical sum $Z = \sum_k^K e^{-\varepsilon_k/T}$ with $\rho_k = e^{-\varepsilon_k/T}$, then the Rényi entropy, within the thermodynamic formalism [30, 31] $S^R_{q=1/T} = \frac{\ln(\sum_k p_k^q)}{q-1}$, can be expressed in terms of free energy through the use of escort distribution: [46]:

$$S^R_{q=1/T} = \frac{F}{T-1}, \ q=1/T \quad (6).$$

In this case, temperature T is considered as a formal parameter (the number of topics/clusters), which can be changed during the computing experiment. In turn, the Tsallis entropy, written in the form of $S^{Ts}_{q=1/T} = \frac{1-\sum_k p_k^q}{q-1}$, can also be expressed in terms of the Rényi entropy:

$$S^{Ts}_q = \frac{e^{(q-1) \cdot S^R_q} - 1}{q-1} \quad (7),$$

and, consequently, in terms of free energy [46]. Thus, the variation of parameter $q=1/T$ also allows us to investigate the behavior of Tsallis entropy in topic modeling. It should be noted that with this approach, the entropy divergence is achieved at q=1, that is, the information obtained in topic modeling for one topic is zero. On the other hand, as T→∞, we get a uniform probability distribution of words by topics, which also correspond to the maximum entropy or minimum information.

## 4. Numerical investigation of the application of q-deformed entropy to determine the number of topics

### 4.1. Datasets

In the framework of this study, computing experiments were performed on the following datasets.

- The well-known English-language dataset '20newsgroups' [47]: 15,404 news texts; 50,948 unique words. According to the description of the dataset, data is organized into 20 different newsgroups, each corresponding to a different topic. As some of the newsgroups are very closely related to each other, the actual number of topics, according to the authors, corresponds to approximately 15. When conducting topic modeling on this dataset, the number of topics varied in the range: T=[2; 120] in increments of 2 topics.

- All posts of the top 2,000 bloggers of the Russian-language section of social network LiveJournal for January-April 2014: 101,481 posts; 172,939 unique words. This dataset contains a mixture of short conversational messages and long posts in a journalistic style. The number of topics varied in the range: T=[2; 330] in increments of 2 topics.

The choice of these datasets was due to the following reasons. First of all, they are datasets in different languages. Accordingly, our computational experiments show the applicability of the entropy approach to collections in different languages, and also show the common model features inherent in different languages. Secondly, since larger collections (LJ) usually contain a greater quantity of topics, that is, they have great variety, it is logical to assume that they have more local entropy minima requiring separate attention. Thirdly, various clustering models were tested in the English-language collection [48], which makes it possible to compare the results of topic modeling with the results of cluster analysis.

**4.2. Experimental design.**

We studied the four topic models listed below from the point of view of the behavior of Rényi and Tsallis entropies as a function of the number of topics, using the following software implementations:

A. pLSA (E-M algorithm) – BigARTM (http://bigartm.org/).
B. VLDA (E-M algorithm) – Latent Dirichlet Allocation package, http://chasen.org/~daiti-m/dist/lda/.
C. LDA GS (Gibbs sampling) – GibbsLDA++ (http://gibbslda.sourceforge.net/).
D. GLDA (Gibbs sampling) (https://linis.hse.ru/en/soft-linis)

All source codes were reduced to one software tool 'TopicMiner' (https://linis.hse.ru/en/soft-linis/) as a set of dynamic link libraries (dll). Thus, computational experiments were carried out on collections that were processed in exactly the same way.

In each computational experiment, for each model, the number of microstates whose probabilities were greater than the preassigned value $P_{nt}>1/N$ was measured. Further, on the basis of formula 4, the density-of-states function was calculated from the number of topics. Also, internal energy, entropy and free energy were calculated for each topic solution in accordance with formula 5. On the basis of the free energy, Rényi and Tsallis entropies were calculated using formulas 6 and 7 for each topic solution.

**4.3. Discussion of the results of the experiments on the 20 newsgroups dataset**

Figures 1, 2 and 3 show charts of Shannon, Rényi and Tsallis entropies depending on the number of topics for all four models of topic modeling on the 20 newsgroups dataset. Each model was run three times, then the results of the calculation were averaged. The entropy values were calculated on the basis of the averaged values.

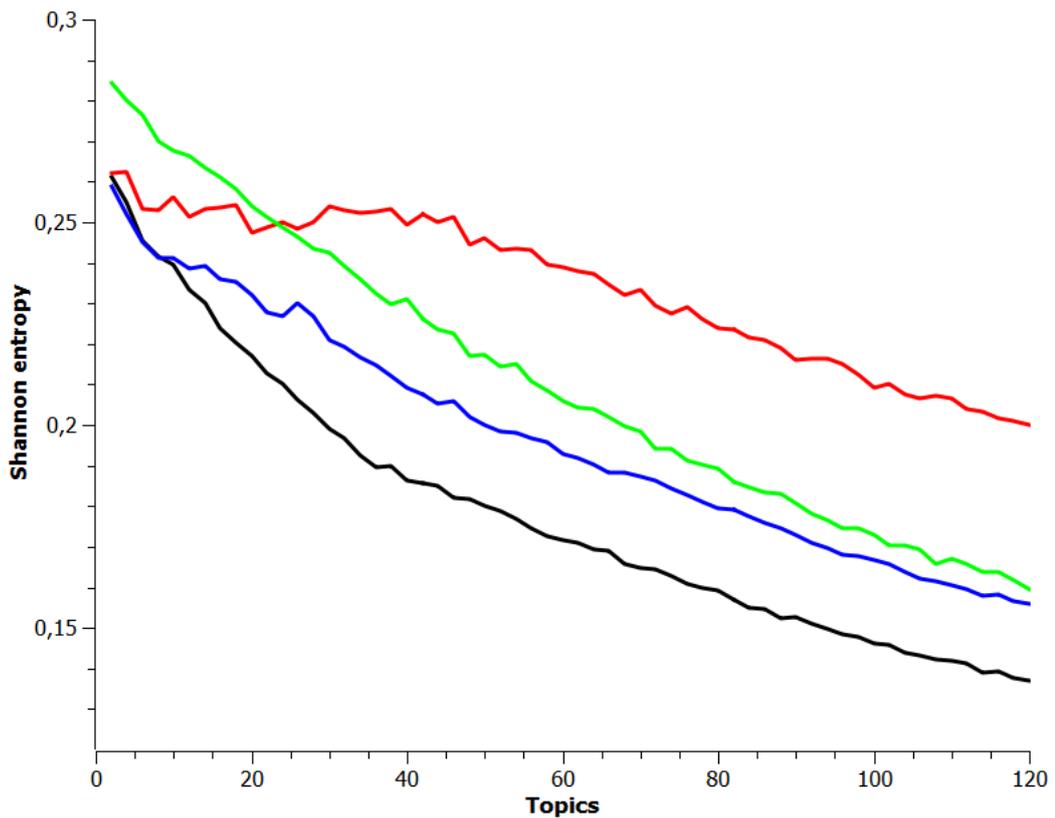

Fig. 1. Shannon entropy as a function of the number of topics on the 20 newsgroups dataset. Black: LDA GS, blue: pLSA; green: VLDA; red: GLDA.

The LDA GS, pLSA and VLDA models give similar curves without a pronounced minimum or maximum, while the GLDA model gives a small maximum in the region of 30-40 topics. The curves in Figure 1 show that the greater the number of topics/clusters, the lower the entropy value and, correspondingly, the greater the value of information. However, this contradicts the actual experimental results, since an increase in the number of topics leads to the probability distribution of words by topics tending to a uniform distribution, which should correspond to an increase in entropy. This means that Shannon entropy is not suitable for analyzing a complex information system, which means that perplexity, the most commonly used quality measure of topic modeling, is not suitable either.

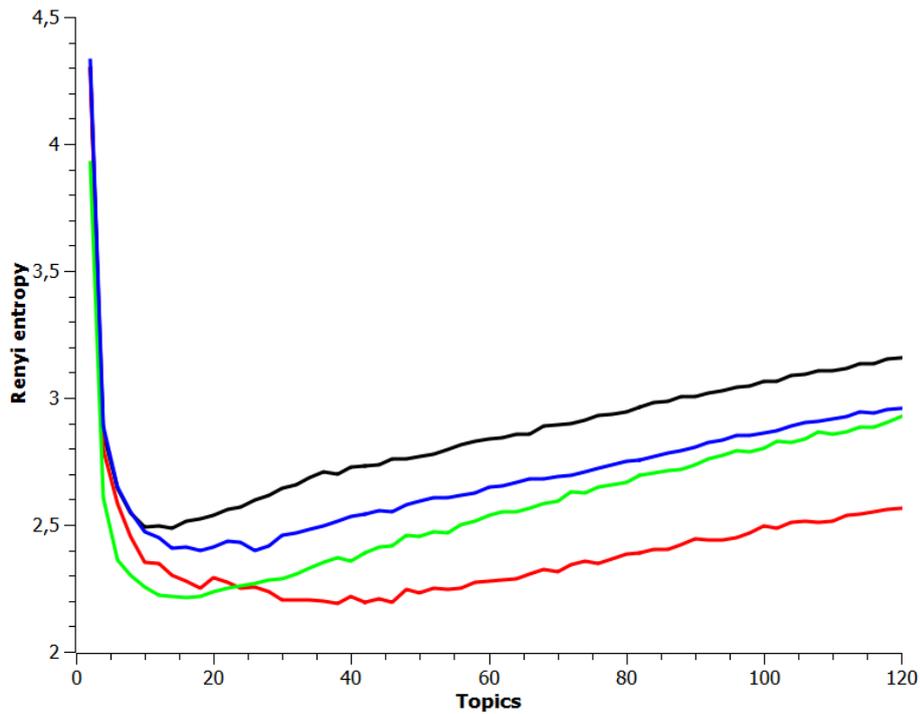

Fig. 2. Rényi entropy as a function of the number of topics on the 20 newsgroups dataset. Black: LDA GS, blue: pLSA; green: VLDA; red: GLDA.

The Rényi entropy, in contrast to the Shannon entropy, has a global minimum, and shows the correct results on the boundary values of the number of topics. For T=1, the entropy should give a maximum, because topic modeling, just like any other cluster algorithm, does not give the distribution of clusters, so information about the cluster distribution is zero. At the same time, as noted above, an excessive increase in the number of clusters/topics (i.e., T→∞) leads to a uniform distribution of each word by topics, which also corresponds to an increase in entropy or a decrease in information. However, different models give slightly different minimum Rényi entropy positions, and different depths for this minimum. In order to determine which of these models gives a more accurate result, it is necessary to compare the results of topic modeling with alternative methods of determining the number of topics in the same collection. The authors of [48] tested a number of clustering algorithms on the '20 newsgroup dataset' and showed that the optimal number of clusters varies from 15 to 20 clusters for different cluster algorithms, which also corresponds to the description of the dataset by its creators.

The LDA GS and VLDA models also show the optimal number of topics in the order of 15, the pLSA model shows 20, while the VLDA shows the deepest minimum. However, the GLDA gives a significant difference, almost doubling the number of topics compared to other models, as well as to alternative methods of determining the number of topics. Thus, it can be said that the procedure for regularizing the GLDA model on the one hand provides almost 100% stability [21], but on the other hand it leads to a shift in the minimum Rényi entropy relative to the correct value.

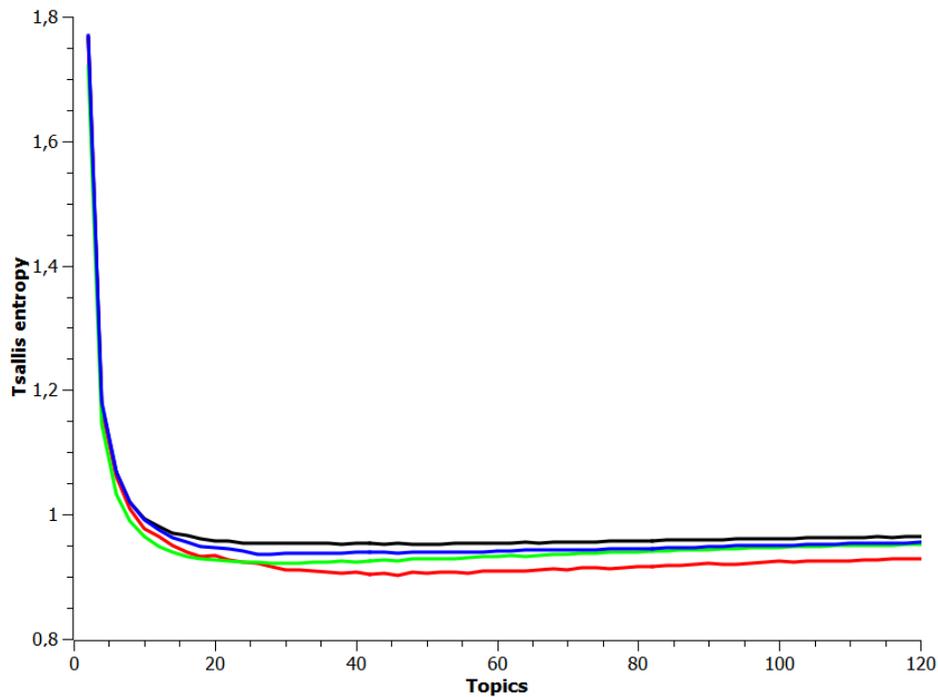

Fig. 3. Tsallis entropy as a function of the number of topics in the 20 newsgroups dataset. Black: LDA GS, blue: pLSA; green: VLDA; red: GLDA.

As can be seen from Fig. 3, the Tsallis entropy curves have significantly less pronounced minima, which makes it difficult to determine the optimal number of topics using them. In this case, the Tsallis entropy, like the Rényi entropy, gives the maximum values at the boundaries of interval $[1; \infty]$. All models, with the exception of the GLDA model, give a minimum in the region of 20 topics, which corresponds to the values obtained by the alternative method.

**4.4. Discussion of the results of the experiments on the LJ dataset.**

A large number of documents in the collection can lead to the appearance of additional local minima, which should also be investigated. Moreover, these additional local minima can be of the greatest interest for researchers of the content of texts, because clustering into 15-20 topics gives topics of too general a nature (such as sports, politics and art), the presence of which in the news flow is evident without research. It can be assumed that solutions that divide global topics into more specific but not excessively fractional topics (for example, "politics in the Middle East" and "European politics") correspond to local minima of nonextensive entropy. In order to verify the presence of such minima, the Rényi and Tsallis entropies were calculated in this paper for a large collection, the LJ dataset. The Rényi and Tsallis entropy curves are shown in Figures 4 and 5.

First of all, it should be noted that the models based on the EM-algorithm show a marked difference from the models based on Gibbs sampling for the LJ dataset. The LDA GS model demonstrates the presence of strong jumps of Rényi entropy, which are associated with significant fluctuations in the density distribution function. However, the VLDA and pLSA models do not see these jumps. Fluctuations in the density distribution in the Gibbs sampling models cannot be explained by the features of the sampling procedure, since in study [42] research was conducted on the same dataset, in which the LDA GS model was run three times for each topic, and the topics varied in increments of 1 in the range of [105-120], and in increments of 10 in the range of [120-600]. The jump in the region [110-120] was observed in all runs of the model. Thus, the models based on the Gibbs sampling are more sensitive than the other models.

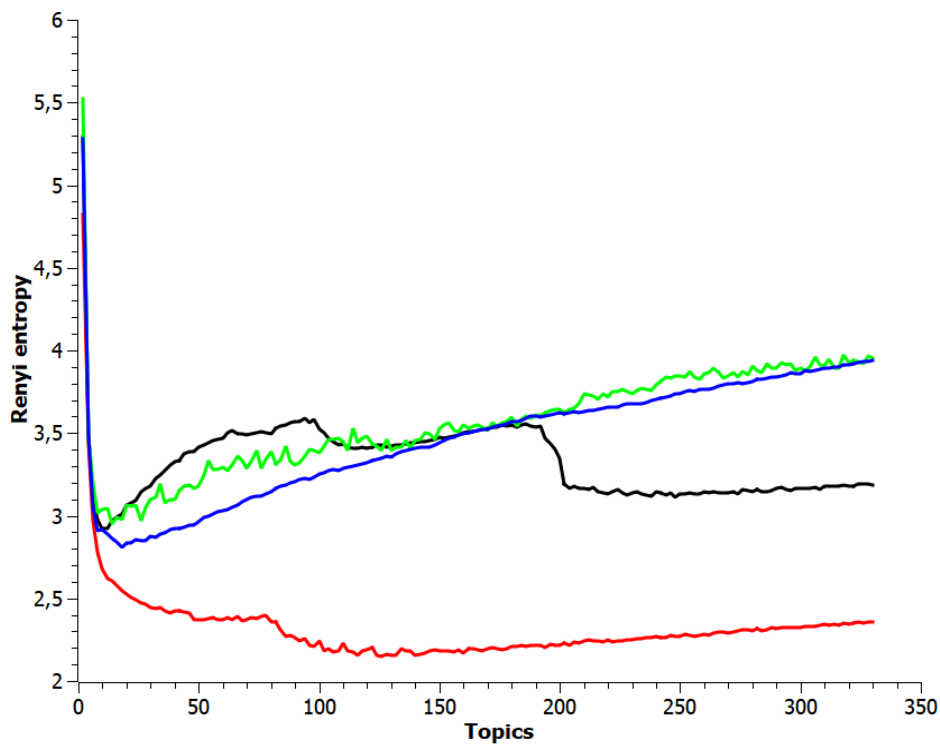

Fig. 4. Rényi entropy as a function of the number of topics in the LJ dataset. Black: LDA GS, blue: pLSA; green: VLDA; red: GLDA

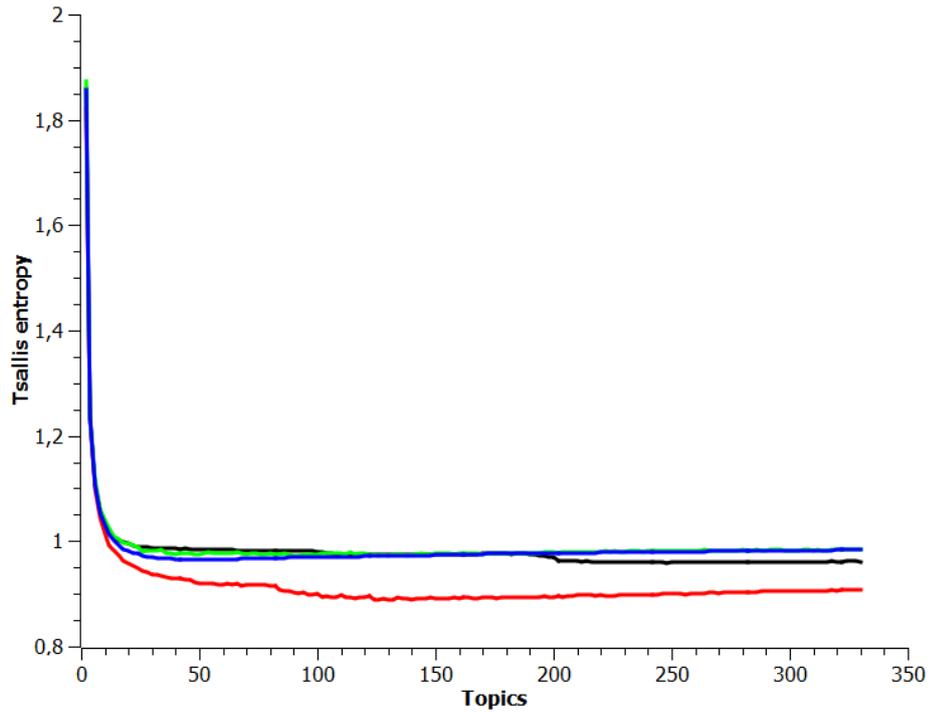

Fig. 5. Tsallis entropy as a function of the number of topics in the LJ dataset. Black: LDA GS, blue: pLSA; green: VLDA; red: GLDA.

The Tsallis entropy calculated on the LDA GS model also shows a jump in the topic region [110-120] and in the topic region [190-200]; however, the amplitude of the jump is much lower. This is due to the fact that Tsallis entropy is more stable from the point of view of Lesche [49, 10]. In this case, the lack of Lesche stability in Rényi entropy is useful from the point of view of revealing local minima.

## 5. Numerical experiments of semantic stability in topic models.

Since, as indicated in Section 2.2., text systems do not have the property of indistinguishability of particles, in investigating their behavior, it is necessary to check whether the word distribution is reproducible from the semantic point of view when the parameter T (the number of topics) changes, that is, how much the composition of words describing topics with high probability is T-invariant.

In this paper, T-invariance in topic models was measured using the Jaccard index [45] by the formula: $J_k=a/(a+b-c)$, where a is the set of the most probabilistic words in topic solution $T_1$ and absent in solution $T_2$, b is the set of words in solution $T_2$ and absent in solution $T_1$, and c is the set of words common for solutions $T_1$ and $T_2$. The coefficient is 1 in the case of complete coincidence of the two sets and is equal to 0 if the sets are completely different. When calculating the coefficient, words with high probability were used, that is, when the probability of the word was $P_w>1/N$, where N is the number of unique words in the collection of documents.

To determine the effect of the number of topics T on the total composition of top words, T varied in increments of 2 in the range from 2 to 120 on the 20 newsgroups dataset, and from 2 to 330 on the LJ dataset. Then a pairwise comparison of each topic solution was made with all the other solutions. As a result of the calculation, a Jaccard index matrix was generated. Each cell of the matrix contains the Jaccard index $J_{t1,t2}$, calculated between the lists of top words of the two solutions for which parameter T takes the values of t1 and t2. Since such a matrix is symmetric with respect to the diagonal elements, the coefficients were calculated for only half the matrix. The list of top words was formed on the basis of words with probabilities greater than 1/N, where N is the size of the dictionary of the collection.

Figures 6 and 7 show the Jaccard diagonal coefficients curves according to the LDA GS and VLDA models for the Russian-language dataset. We do not quote the Jaccard coefficients for the English-language dataset since all models showed approximately the same values in the order of 0.999 for all solution pairs.

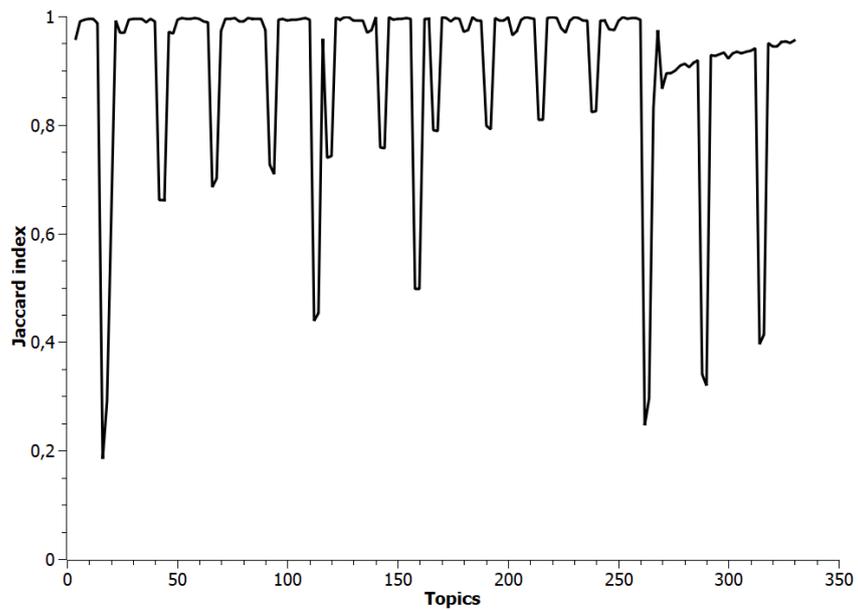

Fig. 6. The diagonal curve of the Jaccard index value for the LDA GS models

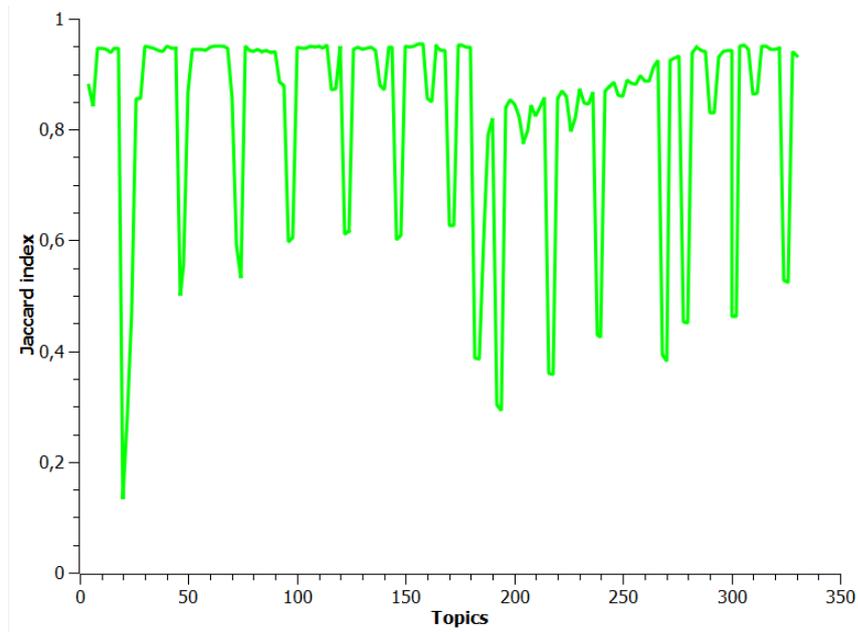

Fig. 7. The diagonal curve of the Jaccard index value for the VLDA models.

Figures 6 and 7 show that, firstly, different models have a similar quasiperiodic semantic structure. This means that when the number of topics T changes, the lists of the most probabilistic words are similar to each other for a different number of topics, but at the same time solutions periodically occur that are very different to the lists of top words from neighbor solutions. Secondly, however, the models demonstrate slightly different behavior. LDA GS shows that in the region of 260 topics, the Jaccard coefficient falls in the stability zones to 0.8, while the VLDA shows that such a fall begins with 190 topics.

Figures 8 and 9 show the 'heat maps' of the Jaccard index for the two models.

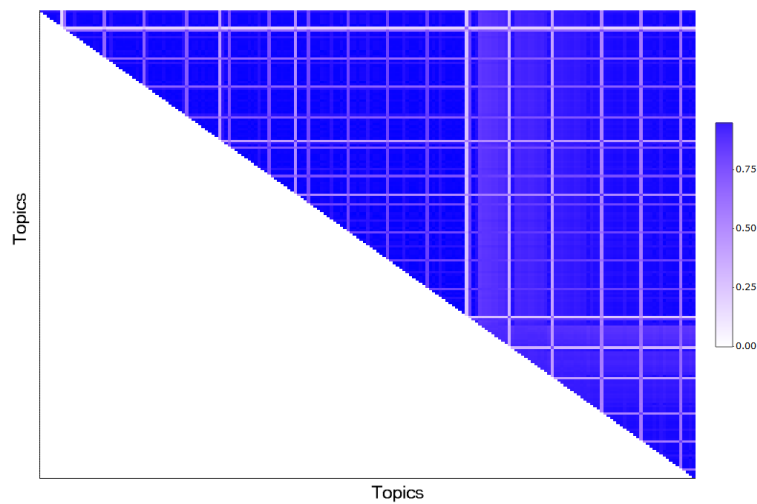

Fig 8. Heat map of Jaccard index distribution for LDA GS models.

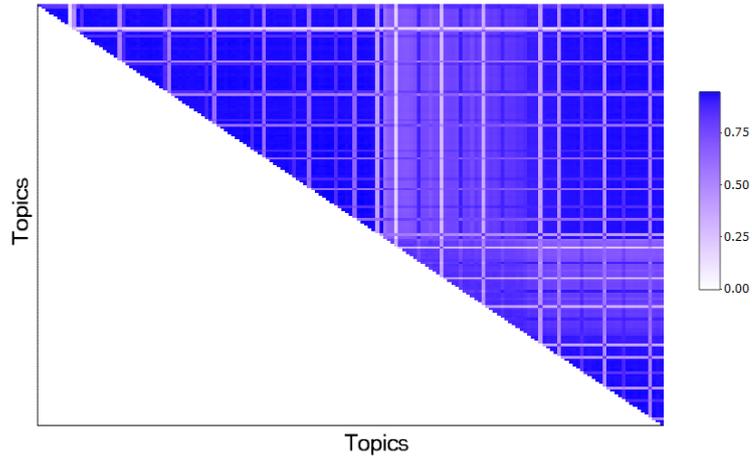

Fig 9. Heat map of Jaccard index distribution for VLDA models.

Figures 8 and 9 also show that models of both types demonstrate the existence of T-invariance zones. At the same time, there exist zones with a high level of coefficient $J_{t1,t2} \cong 0.9$ and zones with a lower level $J_{t1,t2}=0.5$. Moreover, the distributions of Jaccard indices shown in Figures 8 and 9 demonstrate the presence of two quasiperiodic structures that overlap one another. Perhaps the experimentally discovered T-invariance is described by several parameters. Analysis of these parameters is an extremely important task for future studies. Thus, when selecting the optimal number of topics in the topic models, one should choose not only the global and local minima of Rényi or Tsallis entropy, but also avoid zones where the T-invariance principle is violated.

**Conclusion**

In this paper, we formulated an entropy approach to the analysis of the behavior of complex text systems, which allows determining the optimal number of topics in topic models. The study proposed and justified for the first time the application of nonextensive entropy as a function of the number of topics having an extremum, which fundamentally distinguishes it from previously used monotonically decreasing functions (such as Shannon entropy or perplexity) that made it impossible to determine the threshold after which the increase in the number of topics becomes useless or even harmful. Specifically, the search for the optimal number of topics in this paper is based on minimizing Rényi and Tsallis entropies and taking into account T-invariance zones, with Rényi entropy giving the most pronounced minima, which is more convenient for researchers. Both entropies successfully indicate a marked deterioration in the model with an increase in the number of topics in the interval from the optimum to infinity, which corresponds to empirical knowledge and is not captured by the traditionally used monotone functions.

In addition, it is shown that topic models both on the basis of Gibbs sampling and on the basis of EM-algorithms give similar results in the area of the global minimum of nonextensive entropy. However, models based on Gibbs sampling show additional local minima that may be of interest for a comprehensive analysis of large text data in the social sciences.

Finally, the work demonstrates for the first time the existence of two quasiperiodic semantic structures describing the dependence of the change of the total composition of top words of all topics on the number of topics. These structures are not yet included in the theoretical entropy model, but it is already clear that they are essential for determining the optimal number of topics. Further development of the proposed theoretical approach can be obtained by including the two-parameter Sharma-Mittal entropy in this model, which generalizes the entropies of Rényi, Tsallis and Kaniadakis, and possibly will explain the observed quasiperiodic effect. The study was implemented in the framework of the Basic Research Program at the National Research University Higher School of Economics (HSE) in 2017.

**References**

1. Roberts, Margaret E, Brandon M Stewart, and Dustin Tingley. 2016. "Navigating the Local Modes of Big Data: The Case of Topic Models." In *Computational Social Science: Discovery and Prediction*, edited by R Michael Alvarez, 51–97. Cambridge University Press. doi:DOI: 10.1017/CBO9781316257340.004.
2. Brockmann, D., L. Hufnagel, and T. Geisel. 2006. "The Scaling Laws of Human Travel." *Nature* 439 (7075). Nature Publishing Group: 462–65. doi:10.1038/nature04292.
3. Song, C., Koren, T., Wang, P., Barabási, A.L., 2010. Modelling the scaling properties of human mobility. Nature Physics 6, 818–823. doi:10.1038/nphys1760
4. Gleiser, Pablo M., Francisco A. Tamarit, and Sergio A. Cannas. 2000. "Self-Organized Criticality in a Model of Biological Evolution with Long-Range Interactions." *Physica A: Statistical Mechanics and Its Applications* 275 (1). Elsevier Science Publishers B.V.: 272–80. doi:10.1016/S0378-4371(99)00425-2.
5. Friston, K., M. Levin, B. Sengupta, and G. Pezzulo. 2015. "Knowing One's Place: A Free-Energy Approach to Pattern Regulation." *Journal of The Royal Society Interface* 12 (105): 20141383–20141383. doi:10.1098/rsif.2014.1383.
6. Borland, Lisa. 2002. "Option Pricing Formulas Based on a Non-Gaussian Stock Price Model." *Physical Review Letters* 89 (9). doi:10.1103/PhysRevLett.89.098701.
7. Mantegna, Rosario N., H. Eugene Stanley, and Neil A. Chriss. 2000. "*An Introduction to Econophysics: Correlations and Complexity in Finance.*" *Physics Today* 53 (12): 70–70. doi:10.1063/1.1341926.
8. Marković, Dimitrije, and Claudius Gros. 2014. "Power Laws and Self-Organized Criticality in Theory and Nature." *Physics Reports*. doi:10.1016/j.physrep.2013.11.002.

9. Adamic, Lada, and Eytan Adar. 2005. "How to Search a Social Network." *Social Networks* 27 (3): 187–203. doi:10.1016/j.socnet.2005.01.007.

10. Tsallis, Constantino. 2009. *Introduction to Nonextensive Statistical Mechanics: Approaching a Complex World*. Introduction to Nonextensive Statistical Mechanics: Approaching a Complex World. Springer New York. doi:10.1007/978-0-387-85359-8.

11. Bashkirov, A. G. 2004. "On Maximum Entropy Principle, Superstatistics, Power-Law Distribution and Renyi Parameter." In *Physica A: Statistical Mechanics and Its Applications*, 340:153–62. doi:10.1016/j.physa.2004.04.002.

12. Beck, Christian. 2009. "Generalised Information and Entropy Measures in Physics." *Contemporary Physics* 50 (4): 495–510. doi:10.1080/00107510902823517.

13. Blei, David M, Andrew Y Ng, and Michael I Jordan. 2003. "Latent Dirichlet Allocation." *J. Mach. Learn. Res.* 3 (March): 993–1022. doi:http://dx.doi.org/10.1162/jmlr.2003.3.4-5.993.

14. Chernyavsky I, T. Alexandrov, P. Maass, S. Nikolenko, 2012, September. A Two-Step Soft Segmentation Procedure for MALDI Imaging Mass Spectrometry Data. In *GCB* (pp. 39-48)

15. Tu, Nguyen Anh, Dong-Luong Dinh, Mostofa Kamal Rasel, and Young-Koo Lee. 2016. "Topic Modeling and Improvement of Image Representation for Large-Scale Image Retrieval." *Information Sciences* 366 (October): 99–120. doi:10.1016/j.ins.2016.05.029.

16. Daud, Ali, Juanzi Li, Lizhu Zhou, and Faqir Muhammad. 2010. "Knowledge Discovery through Directed Probabilistic Topic Models: A Survey." *Frontiers of Computer Science in China*. doi:10.1007/s11704-009-0062-y.

17. Hofmann, Thomas. 1999. "Probabilistic Latent Semantic Indexing." *Proceedings of the 22nd Annual International ACM SIGIR Conference on Research and Development in Information Retrieval*, 50–57. doi:10.1021/ac801303x.

18. Griffiths, T. L., and M. Steyvers. 2004. "Finding Scientific Topics." *Proceedings of the National Academy of Sciences* 101 (Supplement 1): 5228–35. doi:10.1073/pnas.0307752101.

19. Koltcov, Sergei, Olessia Koltsova, and Sergey I. Nikolenko. 2014. "Latent Dirichlet Allocation: Stability and Applications to Studies of User-Generated Content." *Proceedings of the 2014 ACM Conference on Web Science*, 161–65. doi:10.1145/2615569.2615680.

20. Koltcov, S., S.I. Nikolenko, O. Koltsova, V. Filippov, and S. Bodrunova. 2016. *Stable Topic Modeling with Local Density Regularization. Lecture Notes in Computer Science (Including Subseries Lecture Notes in Artificial Intelligence and Lecture Notes in*

*Bioinformatics)*. Vol. 9934 LNCS. doi:10.1007/978-3-319-45982-016.

21. Koltsov S., Nikolenko S. I., Koltsova O. Gibbs Sampler Optimization for Analysis of a Granulated Medium // Technical Physics Letters. 2016. Vol. 8. No. 42. P. 837-839, doi: 10.1134/S1063785016080241

22. Sugar, Catherine, and James Gareth. 2003. "Finding the Number of Clusters in a Data Set : An Information Theoretic Approach." *Journal of the American Statistical Association* 98: 750–63. doi:10.1198/016214503000000666.

23. Mirkin, Boris. 2005. *Clustering for Data Mining - A Data Recovery Approach*. New York. http://www.amazon.com/dp/1584885343.

24. Tibshirani, Robert, Guenther Walther, and Trevor Hastie. 2001. "Estimating the Number of Clusters in a Data Set via the Gap Statistic." *Journal of the Royal Statistical Society: Series B (Statistical Methodology)* 63 (2): 411–23. doi:10.1111/1467-9868.00293.

25. Fujita, André, Daniel Y. Takahashi, and Alexandre G. Patriota. 2014. "A Non-Parametric Method to Estimate the Number of Clusters." *Computational Statistics & Data Analysis* 73: 27–39. doi:10.1016/j.csda.2013.11.012.

26. Milligan, Glenn W., and Martha C. Cooper. 1985. "An Examination of Procedures for Determining the Number of Clusters in a Dataset." *Psychometrika*. doi:10.1007/BF02294245.

27. Cha, Sung-Hyuk. 2008. "Taxonomy of Nominal Type Histogram Distance Measures." *Proceedings of the American Conference on Applied Mathematics*, no. 2: 325–30. http://www.csis.pace.edu/~scha%5Cnhttp://dl.acm.org/citation.cfm?id=1415583.141564.

28. Cheng, Chun-Hung, Ada Waichee Fu, and Yi Zhang. 1999. "Entropy-Based Subspace Clustering for Mining Numerical Data." In *Proceedings of the Fifth ACM SIGKDD International Conference on Knowledge Discovery and Data Mining - KDD '99*, 84–93. New York, New York, USA: ACM Press. doi:10.1145/312129.312199.

29. Aldana-Bobadilla, Edwin, and Angel Kuri-Morales. 2015. "A Clustering Method Based on the Maximum Entropy Principle." *Entropy* 17 (1). MDPI AG: 151–80. doi:10.3390/e17010151.

30. Rose, Kenneth, Eitan Gurewitz, and Geoffrey C. Fox. 1990. "Statistical Mechanics and Phase Transitions in Clustering." *Physical Review Letters* 65 (8): 945–48. doi:10.1103/PhysRevLett.65.945.

31. Haddad, Wassim M., VijaySekhar Chellaboina, and Sergey G. Nersesov. Thermodynamics: A Dynamical Systems Approach. Princeton University Press, 2005.

32. Stephens, Greg J, Thierry Mora, Gašper Tkačik, and William Bialek. 2013. "Statistical Thermodynamics of Natural Images." *Phys Rev Lett* 110 (1): 18701.

33. Gašper Tkačika, Thierry Morab, Olivier Marrec, Dario Amodeid,e, Stephanie E. Palmerd,f, Michael J. Berry, IIe,g, and William Bialek, Thermodynamics and signatures of criticality in a network of neurons, Proceedings of the National Academy of Sciences of the United States of America. 2015 Sep 15; 112(37): 11508–11513, doi: 10.1073/pnas.1514188112.

34. Venkatesan, R C, and A Plastino. 2011. "Deformed Statistics Free Energy Model for Source Separation Using Unsupervised Learning." *Arxiv Preprint*, no. 2 (February): 5. http://arxiv.org/abs/1102.5396.

35. Martins, André F. T., Noah A. Smith, Eric P. Xing, Pedro M. Q. Aguiar, and Mário A. T. Figueiredo. 2009. "Nonextensive Information Theoretic Kernels on Measures." *Journal of Machine Learning Research* 10: 935--975. http://jmlr.csail.mit.edu/papers/v10/martins09a.html.

36. Venkatesan, R. C., and A. Plastino. 2009. "Generalized Statistics Framework for Rate Distortion Theory." *Physica A: Statistical Mechanics and Its Applications* 388 (12): 2337–53. doi:10.1016/j.physa.2009.02.003.

37. Ramírez-Reyes, Abdiel, Alejandro Raúl Hernández-Montoya, Gerardo Herrera-Corral, and Ismael Domínguez-Jiménez. 2016. "Determining the Entropic Index q of Tsallis Entropy in Images through Redundancy." *Entropy* 18 (8). MDPI AG. doi:10.3390/e18080299.

38. Cao, Juan, Tian Xia, Jintao Li, Yongdong Zhang, and Sheng Tang. 2009. "A Density-Based Method for Adaptive LDA Model Selection." *Neurocomputing* 72 (7–9): 1775–81. doi:10.1016/j.neucom.2008.06.011.

39. Arun, R., V. Suresh, C. E.Veni Madhavan, and M. Narasimha Murty. 2010. "On Finding the Natural Number of Topics with Latent Dirichlet Allocation: Some Observations." In *Lecture Notes in Computer Science (Including Subseries Lecture Notes in Artificial Intelligence and Lecture Notes in Bioinformatics)*, 6118 LNAI:391–402. doi:10.1007/978-3-642-13657-3_43.

40. Teh, Yee Whye, Michael I. Jordan, M. Beal, and David Blei. 2006. "Hierarchical Dirichlet Processes." *Journal of the American Statistical …*, 1–41. doi:10.1017/CBO9781107415324.004.

41. Blei, D., T.L. Griffiths, M.I. Jordan, and J.B. Tenenbaum. 2004. "Hierarchical Topic Models and the Nested Chinese Restaurant Process." *Advances in Neural Information Processing Systems* 16: 106. doi:10.1016/0169-023X(89)90004-9.

42. Koltcov S. N., A thermodynamic approach to selecting a number of clusters based on topic modeling, Technical Physics Letters, 43(6), 584-586, doi.org/10.1134/S1063785017060207
43. Prigogine I. & Stengers I., La Nouvelle Alliance, Gallimard: Paris, 1979.
44. Klimontovich Yu L "Problems in the statistical theory of open systems: Criteria for the relative degree of order in self-organization processes" *Sov. Phys. Usp.* **32** 416–433 (1989).
45. Jaccard, Paul (1912), "The distribution of the flora in the alpine zone", New Phytologist, 11: 37–50, doi:10.1111/j.1469-8137.1912.tb05611.x.
46. Beck C., Schlögl F., *Thermodynamics of Chaotic Systems* (Cambridge University Press, Cambridge, 1993).
47. The 20 Newsgroups data set, http://qwone.com/~jason/20Newsgroups/
48. Basu S, I. Davidson, Wagstaff K. Constrained Clustering: Advances in Algorithms, Theory, and Applications. Chapman & Hall, 2008.
49. Lesche B., Instabilities of Renyi entropies, J. Stat. Phys. **27**, 419 (1982).

# SUPPLEMENTARY MATERIAL

## Topic models used in experiments.

### 1. Probabilistic Latent Semantic Analysis (pLSA)

The formulation of the maximum likelihood problem for pLSA [1] is as follows. As it was said in para. 2.1. of the main article, according to the formula of total probability and the hypothesis of conditional independence, we have the following expression:

$$p(w|d) = \sum_{t \in T} p(w|t) p(t|d) = \sum_{t \in T} \phi_{wt} \theta_{td} \quad \text{S1}$$

Replacing p(w|t) and p(t|d) by matrices $\varphi_{wt}$, $\theta_{td}$ and logarithmizing expression S1 to get rid of the product, we obtain function $L(\varphi, \theta)$, in which it is necessary to fit the values of the matrices so that this expression would be maximal.

$$L(\phi, \theta) = \sum_{d \in D} \sum_{w \in d} n_{wd} \ln \sum_{t \in T} \phi_{wt} \theta_{td} \to max \quad \text{S2,}$$

on condition that $\sum_{t \in T} \theta_{td} = 1$ and $\sum_{w \in W} \phi_{wt} = 1$, $\theta_{td} > 0$, $\phi_{wt} > 0$. The search for the maximum value of function $L(\varphi, \theta)$ is carried out using the so-called EM-algorithm. Its work consists of three stages. At the first stage, matrices $\varphi_{wt}$, $\theta_{td}$ are initialized, and the matrices are filled with random numbers in the range of [0; 1], so that the sum of the numbers in each column (in each topic) of matrix $\varphi_{wt}$ is equal to 1. This means that the sum of probabilities for all words in each topic is always a unity. The sum over all topics is equal to the number of topics T. Matrix $\theta_{td}$ is filled in such a way that the probability of each document belonging to all topics is 1. The second and third stages of the calculation are the steps of the EM-algorithm. At the E-step, the current values of parameters $\varphi_{wt}$, $\theta_{td}$ are used to calculate the conditional probabilities p(t|w) and p(t|d) of all the topics t ∈ T for each term w in each document d by means of Bayes' law. At the M-step, matrices $\varphi_{wt}$ and $\theta_{td}$ are calculated on the basis of the conditional probabilities p(t|w) and p(t|d) in the form of frequency estimates of the corresponding conditional probabilities. The value of $\varphi_{wt}$ is proportional to the number of times $n_{wt}$, when the use of word w was associated with topic t. The value of $\theta_{td}$ is proportional to the number of words $n_{dt}$ in document d related to topic t [2]. At the next iteration, the current matrices $\varphi_{wt}$ and $\theta_{td}$ are used to calculate the probability $p(w|d)$ at the E-step. Then, the M-step is restarted and so on. The process is repeated until matrices $\varphi_{wt}$ and $\theta_{td}$ converge.

### 2. Variational Latent Dirichlet allocation

The further development of the pLSA model is the 'Latent Dirichlet allocation' model (LDA), proposed by David Blei [3]. Blei suggested that the columns of matrices $\varphi_{wt}$ and $\theta_{td}$ in the pLSA model are generated by Dirichlet distributions. Each distribution is characterized by its parameter α (for words) and β (for documents). Also this model assumes that the probability of meeting each topic is determined by a multinomial distribution:

$$p(x_d|\theta_d) = \frac{n_d!}{\prod_t n_{td}!} \prod_t \theta_{td}^{n_{td}} \qquad S3.$$

These assumptions essentially simplify the Bayesian inference of the model due to the properties of the conjugacy of the Dirichlet distributions with the multinomial distribution. The pLSA model, taking into account the Dirichlet functions, is transformed into the LDA model:

$$p(\theta, z, w|\alpha, \beta) = p(\theta|\alpha) \prod_{n=1}^{N} p(z_n|\theta) p(w_n|z_n, \beta) \qquad S4.$$

To determine matrices $\varphi_{wt}$ and $\theta_{td}$ in the LDA model for the collection of documents, the variational Bayesian inference is used with the help of the EM-algorithm. The final formulas for calculating the matrices are as follows:

$$\theta_{td} = \frac{n_{td} + \beta_t}{n_d + \beta_0}, \quad \phi_{td} = \frac{n_{wt} + \alpha_w}{n_t + \alpha_0} \qquad S5,$$

where $n_{td}$ is the number showing how many times document d was encountered in topic t, $n_d$ is the total number of documents in the collection, $\beta_t$ is the value of the Dirichlet function parameter for the document in topic t, $n_{wt}$ is the number showing how many times word w was encountered in topic t, and $\alpha_t$ is the value of the Dirichlet function parameter for the word in topic t. Thus, an inherent difference between the LDA model and the pLSA model is that in the Blei model, the distribution of words and topics for each topic is described by Dirichlet functions that are characterized by the parameters $\beta_t$, $\alpha_t$. However, the methods for calculating matrices $\varphi_{wt}$ and $\theta_{td}$ in the pLSA and LDA models are the same.

3. **Latent Dirichlet Allocation with Gibbs Sampling.**

An alternative way to define matrices $\varphi_{wt}$ and $\theta_{td}$ in topic modeling is the approach of Steyvers and Griffiths [4]. They used the Potts model [5] as a basis for developing an algorithm for classifying words and documents by topics. To explain this model, we can draw the following analogy. Each document is represented as a one-dimensional lattice, and each word in the document as a node. A node can be in one of the T states (topic or cluster). The purpose of the study of such a Potts model is to obtain the distribution of nodes (words) over a set of states, that is, by topics or clusters. The difference between the classical Potts model and topic modeling is that there can be a huge number of TM documents (lattices), and the probability of a node-word belonging to a particular topic-state is determined not only by word-by-topic distribution in one document, but also by distributions of topics over a set of documents. Proceeding from this, the probability that word $w_i$ belongs to topic j is determined by the following expression [4]:

$$p(w_i) = \sum_{j=1}^{T} P(w_i|z_i = j) P(z_i = j) \qquad S6,$$

where $z_i$ is a hidden variable (that is, a topic). In addition, Griffiths and Steyvers proposed, in accordance with the Blei model [3], to use one-dimensional symmetric Dirichlet functions. Finally, to simplify the calculations, they assumed that all the functions in matrix $\varphi_{wt}$ are characterized by a single parameter α, and in matrix $\theta_{td}$ by a single parameter β, with α not equal to β. This simplification makes it possible to obtain analytical expressions connecting the probability of a word in a document and counters:

$$p(w_i) = \sum_{j=1}^{T} P(w_i|z_i = j)P(z_i = j) \cong \frac{C_{mj}^{wt}+\beta}{\sum_m C_{mj}^{wt}+V\beta} \frac{C_{dj}^{dt}+\alpha}{\sum_m C_{dj}^{dt}+\alpha T} \quad \text{S7,}$$

and also to obtain a connection between matrices $\phi_{wt}$ and $\theta_{td}$ and the same counters:

$$\theta_{dj} = \frac{C_{dj}^{dt}+\alpha}{\sum_m C_{dj}^{dt}+\alpha T}, \phi_{dj} = \frac{C_{mj}^{wt}+\beta}{\sum_m C_{mj}^{wt}+V\beta} \quad \text{S8,}$$

where the counters: $C_{mj}^{wt}$ is the number showing how many times word **w** was encountered in topic **t**; $C_{dj}^{dt}$ the number showing how many times word w in document d was associated with topic t; $\sum_m C_{mj}^{wt}$ is the number of words associated with topic t; $\sum_m C_{dj}^{dt}$ is the length of the document in words. In this approach, the probabilities of words and documents belonging to a particular topic are presented as mean values, with the model for calculating the mean values being based on the calculation of multidimensional integrals by the Gibbs sampling method.

The calculation algorithm consists of three stages. At the first stage, the matrices and counters are initialized, the number of topics and parameters α and β are specified, and the number of iterations is determined. The counters that determine the initial content of matrices $\phi_{wt}$ and $\theta_{td\ are}$ given in the form of constants. For example, if you assign an equal belonging of each word to each topic or set this belonging to zero, this will correspond to the situation where matrix $\varphi_{wt}$ is filled with a flat distribution in which all elements of the matrix are equal to the value 1/N, where N is the number of unique words in the document collection.

At the second stage in the cycle (the sampling procedure), all documents and words in each document are re-sorted. Each word from the current document is correlated with a topic number that is generated based on formula S7, that is, the probability of the current word's belonging to different topics is calculated, and then the topic number with the maximum probability that is correlated with the current word is selected. Since the counters are initially equal to either zero or a constant, the probability of correlating the topic and the word at the initial stage is determined solely by values 1/T and 1/N. However, after each correlation of a word and a topic, the counters change, so after a sufficiently large number of iterations, the counters accumulate the complete statistics of the collection of documents being examined.

At the third stage, after the end of the sampling, matrices $\varphi_{wt}$ and $\theta_{td}$ are calculated on the basis of the counter values by formulas S8.

## 4. The Granulated LDA (GLDA) Model

This model was developed in [6] in order to overcome the instability of the LDA, that is, its tendency to produce different results for different runs on the same data with the same parameters, which is due to the influence of the initial distribution on the results. The GLDA model differs from the LDA model on the basis of Gibbs sampling by the size of the sampling region. In the second stage of the sampling procedure, the topic number is assigned not to one word, but to a group of words within a region of a given size. Such a modification of sampling is a kind of regularization that is aimed at increasing the stability of the topic modeling [7]. The selection of the GLDA model for experiments in this work is due to the fact that this model provides practically 100% stability of the Gibbs sampling algorithm [7]; thus, it is possible to see how the almost completely deterministic algorithm behaves in experiments.

**References to supplementary material.**

50. Hofmann, Thomas. 1999. "Probabilistic Latent Semantic Indexing." *Proceedings of the 22nd Annual International ACM SIGIR Conference on Research and Development in Information Retrieval*, 50–57. doi:10.1021/ac801303x.
51. Vorontsov, Konstantin, and Anna Potapenko. 2015. "Additive Regularization of Topic Models." *Machine Learning* 101 (1–3). Kluwer Academic Publishers: 303–23. doi:10.1007/s10994-014-5476-6.
52. Blei, D.M., A.Y. Ng, and M.I. Jordan. 2003. "Latent Dirichlet Allocation." *Journal of Machine Learning Research* 3 (4–5): 993–1022. doi:10.1162/jmlr.2003.3.4-5.993.
53. Griffiths, T. L., and M. Steyvers. 2004. "Finding Scientific Topics." *Proceedings of the National Academy of Sciences* 101 (Supplement 1): 5228–35. doi:10.1073/pnas.0307752101.
54. Landau, David P., and Kurt Binder. 2009. *A Guide to Monte Carlo Simulations in Statistical Physics. Cambridge University Press*. Cambridge University Press. doi:10.1017/CBO9780511994944.
55. Koltsov S., Nikolenko S. I., Koltsova O. Gibbs Sampler Optimization for Analysis of a Granulated Medium // Technical Physics Letters. 2016. Vol. 8. No. 42. P. 837-839, doi: 10.1134/S1063785016080241
56. Koltcov, S., S.I. Nikolenko, O. Koltsova, V. Filippov, and S. Bodrunova. 2016. *Stable Topic Modeling with Local Density Regularization. Lecture Notes in Computer Science (Including Subseries Lecture Notes in Artificial Intelligence and Lecture Notes in Bioinformatics)*. Vol. 9934 LNCS. doi:10.1007/978-3-319-45982-016.